\documentclass[a4paper]{llncs}

\newif\ifdraft\drafttrue
\newif\ifinlineref\inlinereffalse
\newif\iffinal\finalfalse
\newif\ifextended\extendedfalse
\newif\ifdotikz\dotikzfalse

\inlinereftrue
\draftfalse
\finaltrue

\def\papertitle{Nested HEX-Programs}

\usepackage{times}
\usepackage[scaled]{helvet}
\usepackage{courier}

\usepackage{latexsym}
\usepackage{amsmath}
\usepackage{amssymb}
\usepackage{url}
\makeatletter
\let\UrlSpecialsOld\UrlSpecials
\def\UrlSpecials{\UrlSpecialsOld\do\/{\Url@slash}\do\_{\Url@underscore}}%
\def\Url@slash{\@ifnextchar/{\mathchar47\kern-.1em}%
   {\kern-.0em\mathchar47\kern-.08em\penalty\UrlBigBreakPenalty}}
\def\Url@underscore{\nfss@text{\leavevmode \kern.06em\vbox{\hrule\@width.3em}}}
\makeatother

\usepackage{color}
\usepackage[vlined,algoruled]{algorithm2e}
\usepackage{a4,a4wide}

\iffinal\else
\fi
\ifdraft
\usepackage{comments}
\else
\newcommand{\comment}[1]{}
\fi

\usepackage{paralist}
\usepackage{enumerate}
\usepackage{booktabs}
\usepackage{alltt}
\usepackage{caption}

\usepackage{graphicx}
\graphicspath{{figures/}}
\usepackage{subfig}

\usepackage{ifpdf}
\ifpdf
\usepackage{microtype}
\fi

\ifdraft
\usepackage{pdfsync}
\usepackage{srcltx}
\else
\fi

\ifdotikz
\usepackage{tikz}
\pgfrealjobname{nestedhex}
\usetikzlibrary{shapes,arrows,backgrounds,chains,%
matrix,patterns,arrows,decorations.pathmorphing,decorations.pathreplacing,%
positioning,fit,calc,decorations.text,shadows%
}
\else
\long\def\beginpgfgraphicnamed#1#2\endpgfgraphicnamed{\includegraphics{#1}}
\fi

\newcommand{\leanparagraph}[1]{\smallskip\noindent\textbf{#1}. }

\newcommand\nop[1]{}

\DeclareMathOperator{\naf}{not}

\newcommand{\amp}[1]{\ensuremath{\text{\textsl{{\&}}}\!\mathit{#1}}}
\newcommand{\ext}[3]{\ensuremath{\amp{#1}[#2](#3)}}
\newcommand{\extfun}[1]{\ensuremath{f_{\text{\sl\&}#1}}}

\newcommand{\extsem}[4]{\ensuremath{f_{\text{\sl\&}#1}(#2,#3,#4)}}

\newcommand\hex{{\sc hex}}
\newcommand\dlvhex{{\sc dlvhex}}
\newcommand{\meld}{MELD}

\newcommand{\dlv}[0]{\texttt{DLV}}
\newcommand{\clasp}[0]{{\sc clasp}}
\newcommand{\smodels}[0]{{\sc smodels}}

\catcode`\@=11 %
\long\def\blank#1{\bl@nk#1@@..\bl@nk}%
\long\def\bl@nk#1#2@#3#4\bl@nk{#3#4}
\catcode`\@=12

\long\def\test#1{\begingroup \toks0{[#1]}%
  \newlinechar`\/\message{/\the\toks0:
  \if\blank{#1}EMPTY\else NOT empty\fi%
}\endgroup}
\newcommand{\mext}[4]{\ensuremath{#1\if\blank{#3}\else[#3]\fi.{#2}\if\blank{#4}\else(#4)\fi}}

\iffinal
\ifpdf
\fi
\fi

\title{\papertitle%
\iffinal%
\thanks{This research has been supported by the Austrian Science
    Fund (FWF) project P20840 and P20841, and by the Vienna Science and Technology
    Fund (WWTF) project ICT 08-020.}
\fi
}

\author{Thomas Eiter \and Thomas Krennwallner \and Christoph Redl}
\institute{
Institut f\"ur Informationssysteme, Technische Universit\"at Wien\\
\iffinal%
Favoritenstra\ss{}e\ 9-11, A-1040 Vienna, Austria\\
\fi
\email{$\{$eiter,tkren,redl$\}$@kr.tuwien.ac.at}
}

\newcommand{\footnoteremember}[2]{%
  \footnote{#2}%
  \newcounter{#1}%
  \setcounter{#1}{\value{footnote}}%
}

\begin{document}

\maketitle

\begin{abstract}
  Answer-Set Programming (ASP) is an established declarative programming
  paradigm. However, classical ASP lacks subprogram calls
  as in procedural programming, and access to external computations
  (like remote procedure calls) in general. The feature is desired for increasing
  modularity and---assuming proper access in place---(meta-)reasoning
  over subprogram results.  While \hex-programs
  extend classical ASP with external source access,
  they do not support calls of (sub-)programs upfront. We present
  \emph{nested \hex-programs}, which extend
  \hex-programs to serve the desired feature, in a user-friendly manner.
   Notably, the answer sets of called sub-programs can be individually accessed.
  This is particularly useful for applications that need to reason over answer sets like belief set
  merging, user-defined aggregate functions, or preferences of answer sets.
\end{abstract}

\section{Introduction \ifdraft(1-1.5 pages)\fi}

Answer-Set Programming, based on~\cite{gelf-lifs-91}, has been
established as an important declarative programming
formalism~\cite{brew-etal-11-asp}.  However, a shortcoming of classical
ASP is the lack of means for modular programming, i.e., dividing
programs into several interacting components. Even though reasoners such
as \dlv{}, \clasp{}, and \dlvhex{} allow to partition programs into
several files, they are still viewed as a single monolithic sets of
rules.
On top of that, passing input to selected (sub-)programs is not possible
upfront.

In procedural programming, the idea of calling subprograms and
processing their output is in permanent use.  Also in functional
programming such modularity is popular. This helps reducing development
time (e.g., by using third-party libraries), the length of source code,
and, last but not least, makes code human-readable. Reading,
understanding, and debugging a typical size application
written in a monolithic program is cumbersome.
Modular extensions of ASP have been
considered~\cite{jotw2009-jair,lpnmr-97} with the aim of building an
overall answer set from program modules; however, multiple results of
subprograms (as typical for ASP) are respected, and no reasoning about
such results is supported.
XASP~\cite{sw2010-xsb} is an \smodels{} interface for XSB-Prolog.  This
system is related to our work, but in this scenario the meta-reasoner is
Prolog and thus different from the semantics of its subprograms, which
are under stable model semantics.  The subprograms are monolithic
programs and cannot make further calls.  This is insufficient for some
applications, e.g., for the \meld{} belief set merging system, which
require hierarchical nesting of arbitrary depth. Adding such nesting to
available approaches is not easily possible and requires to adapt
systems similar to our approach.

\hex-programs~\cite{eite-etal-2005-ijcai} extend ASP with higher-order
atoms, which allow the use of predicate variables, and external atoms,
through which external sources of computation can be accessed.
But \hex-programs do not support modularity and meta-reasoning
directly. In this context, modularity means the encapsulation of
subprograms which interact through well-defined interfaces only, and
meta-reasoning requires reasoning over \emph{sets of} answer sets.
Moreover, in \hex-programs external sources are realized as procedural
C++ functions.  Therefore, as soon as external sources are queried, we
leave the declarative formalism.  However, the generic notion of
external atom, which facilitates a bidirectional data flow between the
logic program and an external source (viewed as abstract Boolean
function), can be utilized to provide these features.

To this end, we present \emph{nested \hex-programs}, which support
(possibly parameterized) \emph{subprogram calls}.
It is the nature of nested hex-programs to have multiple \hex-programs
which reason over the answer sets of each individual subprogram.  This
can be done in a user-friendly way and enables the user to write purely
\emph{declarative} applications consisting of multiple interacting
modules. Notably, call results and answer sets are \emph{objects}\/ that
can be accessed by identifiers and processed in the calling program.
Thus, different from~\cite{jotw2009-jair,lpnmr-97} and related
formalisms, this enables \emph{(meta)-reasoning about the set of answer
  sets} of a program.
In contrast to~\cite{sw2010-xsb}, both the calling and the called
program are in the same formalism. In particular, the calling program
has also a multi-model semantics.
As an important difference to~\cite{aad2011-tocl}, nested \hex-programs
do not require extending the syntax and semantics of the underlying
formalism, which is the \hex-semantics.  The integration is, instead, by
defining some external atoms (which is already possible in ordinary
\hex-programs), making the approach simple and user-friendly for many
applications.  Furthermore, as nested \hex-programs are based on
\hex-programs, they additionally provide access to external sources
other than logic programs. This makes nested \hex-programs a powerful
formalism, which has been implemented using the \dlvhex{} reasoner for
\hex-programs; applications like belief set merging~\cite{ekr2011-padl}
show its potential and usefulness.
\section{\hex-Programs \ifdraft(0.5 pages)\fi}
We briefly recall \hex-programs, which have been introduced
in~\cite{eite-etal-2005-ijcai} as a generalization of (disjunctive)
extended logic programs under the answer set
semantics~\cite{gelf-lifs-91}; for more details and background, we refer
to~\cite{eite-etal-2005-ijcai}.
A \hex-program consists of rules of the form
\begin{equation*}
  a_1\lor\cdots\lor a_n \leftarrow b_1,\ldots, b_m, \naf\, b_{m+1},
  \ldots, \naf\, b_n \ ,
\end{equation*}
where each $a_i$ is a classical literal, i.e., an atom
$p(t_1,\ldots,t_l)$ or a negated atom $\neg p(t_1,\ldots,t_l)$,
and each $b_j$ is either a classical literal or an external atom,
and  $\naf$ is negation by failure (under stable semantics).
An \emph{external atom} is of the form
\begin{equation*}
  \ext{g}{q_1,\dotsc,q_k}{t_1,\dotsc,t_l} \ ,
\end{equation*}
where $g$ is an external predicate name, the $q_i$ are predicate names
or constants, and the $t_j$ are terms.
Informally, the
semantics of an external $g$ is given by a $k+l+1$-ary Boolean
\emph{oracle function} $\extfun{g}$.  The external atom is true relative
to an interpretation $I$ and a grounding substitution $\theta$ iff
$\extsem{g}{I}{q_1,\dotsc,q_k}{t_1 \theta, \dotsc, t_l \theta} = 1$.
Via such atoms, arbitrary (computable) functions can be
included. E.g., built-in functions can be realized via external
atoms, or library functions such as string manipulations, sorting
routines, etc. As external sources need not be on the same machine,
knowledge access across the Web is possible, e.g., belief
set import.
Strictly,~\cite{eite-etal-2005-ijcai} omits classical negation $\neg$
but the extension is routine;
furthermore,~\cite{eite-etal-2005-ijcai} also allows terms for the~$q_i$
and variables for predicate names, which we do not consider.
\begin{example}
\label{ex:ExternalAtom}
Suppose an external knowledge base consists of an RDF file located on
the web at \url{http://.../data.rdf}. Using an external
atom $\ext{rdf}{<\!url\!>}{X,Y,Z}$, we may access all RDF triples $(s,p,o)$
at the URL specified with $<\!url\!>$. To form belief sets of pairs that
drop the third argument from RDF triples, we may use the rule
\begin{equation*}
  bel(X,Y) \leftarrow
  \ext{rdf}{\text{\url{http://.../data.rdf}}}{X,Y,Z} \ .
\end{equation*}
\end{example}

The semantics of \hex-program is given via answer sets, which are sets
of ground literals closed under the rules that satisfy a stability
condition as in~\cite{gelf-lifs-91}; we refer
to~\cite{eite-etal-2005-ijcai} for technical details. The above program
has a single answer set which consists of all literal $bel(c_1,c_2)$
such some RDF triple $(c_1,c_2,c_3)$ occurs at the respective URL.

We use
the \dlvhex{} system from
{\small\url{http://www.kr.tuwien.ac.at/research/systems/dlvhex/}} as a
backend. \dlvhex{} implements (a fragment of) \hex-programs. It provides
a plugin mechanism for external atoms. Besides library atoms, the user
can defined her own atoms, where for evaluation a C++ routine must be
provided.

\section{Nested \hex-Programs \ifdraft(2-3 pages)\fi}

\leanparagraph{Limitations of ASP}
As a simple example demonstrating the limits of ordinary
ASP, assume a program computing the shortest paths between
two (fixed) nodes in a connected graph. The answer sets of this program
then correspond to the shortest paths.
Suppose we are just interested in the \emph{number} of such paths. In a
procedural setting, this is easily computed: if a function returns all
these paths in an array, linked list, or similar data structure, then
counting its elements is trivial.

In ASP, the solution is non-trivial if the given program must not be
modified (e.g., if it is provided by a third party); above, we must
count the answer sets.  Thus, we need to reason on \emph{sets of} answer
sets, which is infeasible inside the program. Means to call the program
at hand and reason about the results of this \emph{``callee''}
(\emph{subprogram}) in the \emph{``calling program''} (\emph{host
  program}) would be useful.  Aiming at a logical counterpart to
procedural function calls, we define a framework which allows to input
facts to the subprogram prior to its execution.
Host and subprograms are decoupled and interact merely by relational
input and output values. To realize this mechanism, we exploit external
atoms, leading to nested \hex-programs.

\leanparagraph{Architecture}
Nested \hex-programs are realized as a plugin for the reasoner
\dlvhex{},\footnoteremember{meldurl}{\url{http://www.kr.tuwien.ac.at/research/systems/dlvhex/meld.html}}
which consists of a
\emph{set of external atoms} and an \emph{answer cache} for the results
of
subprograms (see Fig.~\ref{fig:architecture}).
Technically, the implementation is part of the
belief set merging system \meld{}, %
which is an application on top of a nested \hex-programs core.
This core can be used independently from the rest of the system.

When a subprogram call (corresponding to the evaluation of a special external atom)
is encountered in the host program, the plugin creates
another instance of the reasoner to evaluate the
subprogram. Its result is then stored in the answer cache and identified
with a unique \emph{handle}, which can later be used
to reference the result and access its components (e.g.,
predicate names, literals, arguments) via other special external
atoms.

There are two possible sources for the called subprogram:
\begin{inparaenum}[(1)]
\item\label{demb} either it is \emph{directly embedded} in the host
  program, or
\item\label{ssep} it is \emph{stored in a separate file}.
\end{inparaenum}
In \eqref{demb}, the rules of the subprogram must be represented
within the host program. To this end, they are encoded as string
constants.
An embedded program must not be confused with a subset of the rules of
the host program. Even though it is syntactically part of it, it is
logically separated to allow independent evaluation.
In \eqref{ssep} merely the \emph{path} to the location of the
external program in the file system is given.
Compared to embedded subprograms, code can be reused
without the need to copy,
which is clearly advantageous when the subprogram changes.
We now present concrete external atoms~$\amp{callhex_n}$,
$\amp{callhexfile_n}$, $\amp{answersets}$, $\amp{predicates}$, and
$\amp{arguments}$.
\begin{figure}[t]
  \centering
  \beginpgfgraphicnamed{systemarchitecture}
  \begin{tikzpicture}[%
    start chain,
    node distance=1cm,
    every on chain/.style={join=by ->},
    every join/.style={line width=1.25pt}]
    \matrix (m) [matrix of nodes,
    column sep=7mm,
    row sep=1.5mm,
    nodes={draw, %
      line width=0.7pt,
      anchor=center,
      text centered,
      text width=1.75cm
    },
    bases/.style={
      minimum width=1.6cm,
      text width=1.6cm,
      tape
    },
    doc/.style={
      tape,
      text width=1.5cm,
      minimum width=1.5cm,
      minimum height=9mm
    },
    docs/.style={
      tape,
      copy shadow,
      fill=white,
      text width=1.5cm,
      minimum width=2cm,
      minimum height=11mm
    },
    sets/.style={
      copy shadow,
      fill=white,
      text width=1.25cm,
      minimum width=1.5cm
    },
    subsystem/.style={
      line width=1.25pt,
      text width=1.9cm,
      minimum width=2.2cm,
      minimum height=9mm
    },
    system/.style={
      line width=1.25pt,
      text width=1.9cm,
      minimum width=2.5cm,
      minimum height=9mm
    },
    source/.style={
      shadow xshift=1ex,
      shadow yshift=1ex,
      cylinder,
      fill=white,
      shape border rotate=90,
      shape aspect=.1,
      line width=1.25pt,
      text width=2.5cm,
      minimum width=3cm,
      minimum height=11mm
    }
    ]
    {
      |[doc]| Main \hex-program & |[system]| \dlvhex{}  & |[sets]| Answer Sets \\
      |[docs]| Subprograms  & |[system]| External Atoms & |[source]| Answer Set Cache \\
    };

    { [start chain,every on chain/.style={join}, every join/.style={line width=1.25pt}]
      \path[line width=1pt,dashed,->] (m-1-1) edge node [right] {} (m-1-2);
      \path[line width=1.5pt,<->] (m-1-2) edge node [right] {} (m-2-2);
      \path[line width=1pt,dashed,->] (m-1-2) edge node [right] {} (m-1-3);
      \path[line width=1pt,dashed,->] (m-2-1) edge node [right] {} (m-2-2);
      \path[line width=1pt,dashed,<->] (m-2-2) edge node [right] {} (m-2-3);
   };

  \end{tikzpicture}
  \endpgfgraphicnamed

\caption{System Architecture of Nestex \hex{} (data flow $\dashrightarrow$, control flow $\to$)}
\label{fig:architecture}

\end{figure}

\leanparagraph{External Atoms for Subprogram Handling}
We start with two families of external atoms
\[\ext{\mathit{callhex_n}}{\mathtt{P},p_1,\dotsc,p_n}{H} \quad \text{
  and } \quad
\ext{\mathit{callhexfile_n}}{\mathtt{FN},p_1,\dotsc,p_n}{H}\] that allow
to execute a subprogram given by a string $\mathtt{P}$ respectively in a file
$\mathtt{FN}$; here $n$ is an integer specifying the number of predicate
names $p_i$, $1 \le i \le n$, used to define the input facts.
When evaluating such an external atom relative to an interpretation $I$,
the system adds all facts $p_i(a_1, \dotsc, a_{m_i}) \gets$
over $p_i$ (with arity $m_i$) that are true in $I$
to the specified program,
creates another instance of the reasoner to evaluate it,
and returns a symbolic handle $H$ as result.
For convenience, we do not write $n$ in $\amp{\mathit{callhex_n}}$ and
$\amp{\mathit{callhexfile_n}}$ as it is understood from the usage.
\begin{example}
\label{ex:Calling1}
In the following program, we use two predicates $p_1$ and $p_2$ to
define the input to the subprogram~$\mathtt{sub.hex}$ ($n = 2)$, i.e.,
all atoms over these predicates are added to the subprogram prior to
evaluation.  The call derives a handle $H$ as result.
\begin{equation*}
\begin{array}{rcl}
p_1(x,y) &\leftarrow& \qquad p_2(a) \leftarrow \qquad p_2(b) \leftarrow  \\
\mathit{handle}(H) &\leftarrow& \ext{\mathit{callhexfile}}{\mathtt{sub.hex}, p_1, p_2}{H}
\end{array}
\end{equation*}
\end{example}

A \emph{handle} is a unique integer representing a certain cache entry. %
In the implementation, handles are consecutive numbers starting with $0$.
Hence in the example the unique answer set of the program
is~$\{\mathit{handle}(0)\}$ (neglecting facts).

Formally, given an interpretation $I$,
$\extsem{\mathit{callhexfile}_n}{I}{\mathit{file}, p_1, \dotsc, p_n}{h}
= 1$ iff $h$ is the handle to the result of the program in file
$\mathit{file}$, extended by the facts over predicates $p_1, \dotsc,
p_n$ that are true in $I$.
The formal notion and use of $\amp{callhex_n}$ to call embedded
subprograms is analogous to  $\amp{callhexfile_n}$. %

\begin{example}
\label{ex:Calling3}
Consider the following program:
\begin{equation*}
\begin{array}{rcl}
h_1(H) &\leftarrow& \ext{callhexfile}{\mathtt{sub.hex}}{H} \\
h_2(H) &\leftarrow& \ext{callhexfile}{\mathtt{sub.hex}}{H} \\
h_3(H) &\leftarrow& \ext{callhex}{\mathtt{a \leftarrow.\ b \leftarrow.}}{H}
\end{array}
\end{equation*}
The rules execute the program $\mathtt{sub.hex}$
and the embedded program $P_e=\{a \leftarrow,\ b \leftarrow\}$.
No facts will be added in this example.
The single answer set is $\{h_1(0), h_2(0), h_3(1)\}$ resp.\ $\{h_1(1),
h_2(1), h_3(0)\}$ depending on the order in which the subprograms
are executed (which is irrelevant).  While $h_1(X)$ and~$h_2(X)$ will have the same
value for~$X$, $h_3(Y)$ will be such that $Y\,{\neq}\, X$.
Our
implementation realizes that the
result of the program in $\mathtt{sub.hex}$
is referred to twice but executes it only once;
$P_e$ is (possibly) different from
$\mathtt{sub.hex}$ and thus evaluated separately.
\end{example}

Now we want to determine how many (and subsequently which) answer sets
it has.  For this purpose, we define external atom
$\ext{answersets}{\mathit{PH}}{\mathit{AH}}$ which maps handles
$\mathit{PH}$ to call results to sets of respective answer set handles.
Formally, for an interpretation $I$,
$\extsem{\mathit{answersets}}{I}{h_P}{h_A}=1$ iff $h_A$ is a handle to
an answer set of the program with program handle $h_P$.
\begin{example}
\label{ex:AnswerSets}
The program
\begin{equation*}
\begin{array}{rcl}
\mathit{ash}(\mathit{PH}, \mathit{AH})	&\leftarrow&
\ext{callhex}{\mathtt{a \lor b \leftarrow.}}{\mathit{PH}}, \ext{answersets}{\mathit{PH}}{\mathit{AH}}
\end{array}
\end{equation*}
calls the embedded subprogram $P_e=\{a \vee b \leftarrow.\}$ and retrieves
pairs $(\mathit{PH}, \mathit{PA})$ of handles to its answer sets.
$\amp{callhex}$ returns a handle $PH = 0$ to the result of $P_e$,
which is passed to
$\amp{answersets}$. This atom returns a \emph{set} of answer set
handles ($0$ and~$1$, as $P_e$ has two answer
sets, viz.\ $\{a\}$ and~$\{b\}$).
The overall program has thus the single
answer set $\{\mathit{ash}(0,0), \mathit{ash}(0,1)\}$. As for
each program the answer set handles start with~$0$, only a pair of
program and answer set handles uniquely identifies an answer set.
\end{example}

We now are ready to solve our example
of counting shortest paths from above.
\begin{example}
\label{ex:InsideAnswerSets}
Suppose $\mathtt{paths.hex}$ is the search program and encodes each
shortest path in a separate answer set. Consider the following program:
\begin{equation*}
\begin{array}{rcl}
  \mathit{as}(AH) &\leftarrow&
  \ext{callhexfile}{\mathtt{paths.hex}}{\mathit{PH}},
  \ext{answersets}{\mathit{PH}}{\mathit{AH}} \\
  \mathit{number}(D) &\leftarrow& \mathit{as}(C), D = C + 1, \naf as(D)
\end{array}
\end{equation*}
The second rule computes the first free handle $D$; the latter coincides
with the number of answer sets of~$\mathtt{paths.hex}$ (assuming that
some path between the nodes exists).
\end{example}

At this point we still treat answer sets of subprograms as black
boxes.  We now define an external atom to investigate them.
Given an interpretation $I$, $\extsem{\mathit{predicates}}{I}{h_P, h_A}{p,a}=1$
iff $p$ occurs as an $a$-ary predicate in the answer set identified by
$h_P$ and $h_A$.  Intuitively, the external atom maps pairs of program
and answer set handles to the predicates names with their associated
arities occurring in the accourding answer set.
\begin{example}
\label{ex:ExtractingPredicates}
We illustrate the usage of  $\amp{predicates}$ with the following program:
\begin{equation*}
\begin{array}{rcl}
  \mathit{preds}(P, A) &\leftarrow&
    \ext{callhex}{\mathtt{node(a).\ node(b).\ edge(a, b).}}{\mathit{PH}}, \\
  && \ext{answersets}{\mathit{PH}}{\mathit{AH}}, \ext{predicates}{\mathit{PH}, \mathit{AH}}{P,A}
\end{array}
\end{equation*}
It extracts all predicates (and their arities) occurring in the
answer of the embedded program $P_e$, which specifies a graph.
The single answer set is
$\{\mathit{preds}(\mathit{node},1), \mathit{preds}(\mathit{edge},2)\}$
as the single answer set of $P_e$ has atoms with predicate
$\mathit{node}$ (unary) and $\mathit{edge}$ (binary).
\end{example}

The final step to gather all information from the answer of a subprogram
is to extract the \emph{literals} and their \emph{parameters} occurring in
a certain answer set.  This can be done with external atom
$\amp{arguments}$, which is best demonstrated with an example.
\begin{example}
\label{ex:Arguments}
Consider the following program:
\begin{equation*}
\begin{array}{rcl}
\mathit{h}(\mathit{PH}, \mathit{AH}) &\leftarrow&
\ext{callhex}{\mathtt{node(a).\ node(b).\ node(c).\ edge(a, b).}
  \mathtt{edge(c, a).}}{\mathit{PH}}, \\
 & & \ext{answersets}{\mathit{PH}}{\mathit{AH}} \\
\mathit{edge}(\mathit{W}, \mathit{V})	&\leftarrow& h(\mathit{PH}, \mathit{AH}), \ext{arguments}{\mathit{PH}, \mathit{AH}, \mathtt{edge}}{\mathit{I}, 0, \mathit{V}}, \\
 & & \ext{arguments}{\mathit{PH}, \mathit{AH}, \mathtt{edge}}{\mathit{I}, 1, \mathit{W}} \\
\mathit{node}(\mathit{V}) &\leftarrow& h(\mathit{PH}, \mathit{AH}), \ext{arguments}{\mathit{PH}, \mathit{AH}, \mathtt{node}}{\mathit{I}, 0, \mathit{V}}
\end{array}
\end{equation*}
It extracts the directed graph given by the embedded subprogram $P_e$
and reverses all edges; the single answer set is
$\{h(0,0), \mathit{node}(a), \mathit{node}(b), \mathit{node}(c), \mathit{edge}(b,a),
\mathit{edge}(a,c)\}$. Indeed,
$P_e$ has a single answer set, identified by $PH=0,AH=0$;
via $\amp{arguments}$ we can access in the second resp.\ third rule
the facts over  $\mathit{edge}$ resp.\ $\mathit{node}$ in it, which are
identified by a unique literal id $I$;  the
second output term of~$\amp{arguments}$ is the argument position, and
the third the actual value at this position.
If the predicates of a subprogram were unknown, we can determine them using
$\amp{predicates}$.
\end{example}

To check the sign of a literal, the external atom~$\ext{arguments}{PH,
  AH,Pred}{I, s, \mathit{Sign}}$ supports argument $s$. When
$s=0$, $\amp{arguments}$ will match the sign of the $I$-th
\emph{positive} literal over predicate $\mathit{Pred}$ into
$\mathit{Sign}$, and when $s=1$ it will match the corresponding
classically negated atom.

\section{Applications \ifdraft(1 page)\fi}

\leanparagraph{\meld{}}
The \meld{} system~\cite{ekr2011-padl} deals with
merging multiple
\emph{collections of belief sets}. Roughly, a belief set is
a set of classical
ground literals.
Practical examples of belief sets include explanations in abduction problems,
encodings of decision diagrams, and relational data.
The merging strategy is defined by
tree-shaped \emph{merging plans}, whose leaves are the collections of belief sets to be merged, and whose inner nodes
are \emph{merging operators} (provided by the user).
The structure is akin to syntax trees of terms.

The automatic evaluation of tree-shaped merging plans is based on nested \hex-programs;
it proceeds bottom-up, where every step requires inspection
of the subresults, i.e., accessing the answer sets of subprograms.
Note that for nesting of ASP-programs with arbitrary (finite) depth,
XASP~\cite{sw2010-xsb} is not appropriate.

\leanparagraph{Aggregate Functions}
Nested programs can also emulate aggregate
functions~\cite{flp2011-ai} (e.g., sum, count, max) where the (user-defined) host program
computes the function given the result of a subprogram. This can
be generalized to aggregates over \emph{multiple} answer sets of the
subprogram; e.g., to answer set
counting, or to find the minimum/maximum of some predicate over
all answer sets (which may be exploited for global optimization).

\leanparagraph{Generalized Quantifiers}
Nested \hex-programs make the implementation of brave and cautious
reasoning for query answering tasks very easy, even if the backend
reasoner only supports answer set enumeration.  Furthermore, extended
and user-defined types of query answers (cf.~\cite{lpnmr-97}) are definable in a very
user-friendly way, e.g., majority decisions (at least half of the answer
sets support a query), or minimum and/or maximum number based decisions (qualified number
restrictions).

\leanparagraph{Preferences}
Answer sets as accessible objects can be easily compared wrt.
user-defined preference rules, and used for filtering as well as
ranking results (cf.~\cite{DBLP:journals/ci/DelgrandeSTW04}):
a host program selects appropriate candidates produced by a subprogram,
using preference rules. The latter  can be elegantly
implemented as ordinary integrity constraints (for filtering), or as
rules (possibly involving further external calls) to derive a rank.
A popular application are online shops, where the past consumer behavior
is frequently used to filter or sort search results.  Doing the
search via an ASP program which delivers the matches in answer sets, a
host program can reason about them and act as a filter or ranking
algorithm.

\section{Conclusion \ifdraft(0.5-1 pages)\fi}

To overcome limitations of classical ASP regarding subprograms and
reasoning about their possible outcomes, we briefly presented
\emph{nested \hex-programs}, which realize subprogram calls via special
external atoms of \hex-programs; besides modularity, a plus for
readability and program reusability, they allow for reasoning over
\emph{multiple} answer sets (of subprograms). An prototype
implementation on top of \dlvhex{} is available.
Related to this is the work on macros in~\cite{bdt2006}, which allow to
call macros in logic programs.

The possibility to access answer sets in a host
program, in combination with access to other external computations,
makes nested \hex-programs a powerful tool for a number of applications.
In particular, libraries and user-defined functions can be incorporated
into programs easily.  As an interesting aspect is that dynamic program
assembly (using a suitable string library) and execution are possible, which
other approaches to modular ASP programming do not offer. Exploring
this remains for future work.

\ifinlineref

\vspace*{-.25\baselineskip}

\else
\bibliographystyle{splncs03}
\bibliography{nestedhex-bib}
\fi

\end{document}